\documentclass[3p,times]{elsarticle}

\usepackage{ecrc}

\volume{}

\firstpage{1}

\journalname{}

\runauth{Hussain Alasmawi \& Leanne Bricker \& Mohammad Yaqub}

\jid{}

\jnltitlelogo{}

\usepackage{amssymb}
\usepackage{adjustbox}
\usepackage{url}
\usepackage{amsmath}
\usepackage{multicol}
\usepackage{setspace}
\usepackage{lineno}
\usepackage{graphicx}
\usepackage{subcaption} 
\usepackage{xcolor}
\definecolor{red}{rgb}{0,0,0}

\usepackage[figuresright]{rotating}

\makeatother

\begin{document}

\begin{frontmatter}



\dochead{}

\title{FUSC: Fetal Ultrasound Semantic Clustering of Second Trimester Scans Using Deep Self-supervised Learning} 


\author[1]{Hussain Alasmawi \corref{cor1}}
\author[2]{Leanne Bricker}
\author[1]{Mohammad Yaqub}
\address[1]{Mohamed bin Zayed University of Artificial Intelligence, Abu Dhabi, United Arab Emirates}
\address[2]{Abu Dhabi Health Services Company (SEHA), Abu Dhabi, United Arab Emirates}
\cortext[cor1]{Corresponding Author: Mohamed bin Zayed University of Artificial Intelligence, Abu Dhabi, United Arab Emirates.
\\
E-mail address: hussain.alasmawi@mbzuai.ac.ae}

\begin{abstract}
{\color{red}
Objective: This study aims to address the challenges posed by the manual labeling of fetal ultrasound images by introducing an unsupervised approach, the Fetal Ultrasound Semantic Clustering (FUSC) method. The primary objective is to automatically cluster a large volume of ultrasound images into various fetal views, reducing or eliminating the need for labor-intensive manual labeling.
\\
Methods: The FUSC method is developed utilizing a substantial dataset comprising 88,063 images. The methodology involves an unsupervised clustering approach to categorize ultrasound images into diverse fetal views. The method's effectiveness is further evaluated on an additional, unseen dataset consisting of 8,187 images. The evaluation includes assessing the clustering purity, and the entire process is detailed to provide insights into the method's performance.
\\
Results: The FUSC method demonstrates notable success, achieving over 92\% clustering purity on the evaluation dataset of 8,187 images. The results signify the feasibility of automatically clustering fetal ultrasound images without relying on manual labeling. The study showcases the potential of this approach in handling the large volume of ultrasound scans encountered in clinical practice, with implications for improving efficiency and accuracy in fetal ultrasound imaging.
\\
Conclusion: The findings of this investigation suggest that the FUSC method holds significant promise for the field of fetal ultrasound imaging. By automating the clustering of ultrasound images, this approach has the potential to reduce the manual labeling burden, making the process more efficient. The results pave the way for advanced automated labeling solutions, contributing to the enhancement of clinical practices in fetal ultrasound imaging. Our code is available at} \url{https://github.com/BioMedIA-MBZUAI/FUSC} 
\end{abstract}


\end{frontmatter}


\section*{Introduction}
\label{intro}
Data labeling is critical for training supervised learning models in medical imaging. Due to its exceptional performance, deep learning (DL) has become the preferred machine learning approach, primarily through supervised methods. However, DL needs a large labeled dataset which requires clinical expertise and is typically time-consuming and resource-intensive. With the growing demand to automate various medical imaging tasks, it is necessary to explore ways that have the potential to reduce costs and improve the efficiency of data labeling.
Self-supervised learning (SSL) is a powerful technique for learning image feature representations without labels. This is achieved using pre-designed pretext tasks, such as inpainting patches \cite{pathak2016context}, rotation prediction \cite{gidaris2018unsupervised}, contrastive learning \cite{chen2020simple,he2020momentum}, and non-contrastive learning \cite{caron2021emerging,grill2020bootstrap}, which do not require labeled data to train the model but instead rely on a proxy task. SSL has been shown to improve fetal ultrasound models for both videos \cite{jiao2020self}, and images \cite{chen2019self}. However, it is essential to note that SSL is typically used as the first stage of a two-stage pipeline. The second stage involves fine-tuning the network in a fully-supervised manner on all or a subset of the labeled data. This means a labeled dataset is still needed to train such models. In addition, as much as SSL methods have demonstrated impressive results in the natural image analysis problems \cite{zhou2021ibot,zhou2022mugs}, it is arguably not as helpful in assessing some medical image applications \cite{zhang2022dive}.
Clustering in DL aims to extract a low-dimensional embedding and group semantically similar images without labels. This can provide a semi-automated labeling tool by creating high-purity clusters. Various methods have been developed for clustering natural images, such as learning a pretext task and clustering at the same time \citep{caron2018deep,ji2019invariant}, and training autoencoders \citep{DEC} or generative models \citep{ji2021decoder} followed by clustering the learned low-dimensional representation. Despite the effectiveness of these methods, they face various issues, such as cluster degeneracy, where samples from different categories are grouped into a single cluster or two or more cluster centers are identical. 
Also, these approaches may not be suitable for medical imaging since they rely on the existence of highly discriminative detailed features that may not be easily extractable from some medical imaging data compared with natural imaging data.
Clustering methods in medical imaging literature are limited compared to natural imaging, highlighting the low attention given to this approach in the medical imaging community. The study by Kart et al. \citep{kart2021deepmcat} applied clustering to cardiac MR images by modifying the DeepCluster framework \citep{caron2018deep} and achieved high-performance results. This approach has shown excellent performance on clustering MRI views. However, the problem is well-defined, and the quality of images is considerably higher than in many other medical image applications, especially ultrasound imaging.  Additionally, Mittal et al. \citep{mittal2021new} introduced a new variation of the gravitational search algorithm for clustering COVID-19 images. Dadoun et al. \citep{dadoun2022deep} employed the idea in \citep{huang2020deep} to cluster ultrasound images of abdominal organs as a pretext task for multi-label classification. Huang and Cui \citep{huang2022breast} utilized clustering to distinguish two groups (benign and malignant tumors) in breast ultrasound images. These methods have either been applied on a small number of clusters or cases with high quality medical image modalities, e.g., MRI, with a minimal amount of noise. 
This study presents FUSC, a DL clustering approach based on extracting important low-dimensional features using SSL, then training a cluster head to disentangle the images into different clusters. FUSC enforces images with a similar appearance to have a closer latent representation in the embedding space. The main contributions of this work are: 
\begin{itemize}
\item proposing an unsupervised clustering method for the challenging task of fetal ultrasound view disentanglement, which is, to our knowledge, the first work in this domain;
\item evaluating the model on a large dataset and demonstrating model generalizability on an additional unseen dataset which includes out-of-distribution classes and images captured from different machines (distribution shift) and achieving a new state-of-the-art performance; 
\item introducing an entropy-based loss in conjunction with the clustering loss to help reduce the effect of having a highly imbalanced dataset during training.
\end{itemize}


\subsection*{Related Work}

In recent years, DL, specifically convolutional neural networks, have played an increasingly important role in analyzing fetal ultrasound images. This has led to the development of a vast body of literature to support healthcare professionals with clinical decisions. The use of DL has allowed for the automation of various tasks in fetal ultrasound imaging, mainly on image classification, anomaly detection, and biomarker measurement.

For the image classification task, \cite{yu2017deep} fine-tuned a shallow classification CNN pre-trained on ImageNet to detect fetal standard plane, and the method was tested on 2,418 images, reaching an AUC, accuracy, precision, recall, and F1-score of 99\%, 96\%, 96\%, 97\%, and 97\% respectively. In another work, \cite{qu2020standard} used a shallow classification CNN to automatically identify fetal standard planes in 19,142 images, with accuracy, precision, recall, and F1-score of 93\%, 93\%, 92\%, and 93\%, respectively.

In \cite{burgos2020evaluation}, the authors compared state-of-the-art CNNs for classifying six different fetal planes and performed testing on a dataset of 5,271 images from 896 subjects. The best-performing network was DenseNet-169, with top-1 error, top-3 error, and average class accuracy of 6.20\%, 0.27\%, and 93.6\%, respectively. In \cite{kong2018automatic}, a dense network was used to detect four fetal standard planes and was tested on 5,678 ultrasound images, with precision, recall, and F1-score of 98\%, 98\%, and 98\%, respectively. The work in \cite{liang2019sprnet} proposed an automatic fetal standard plane classification based on DenseNet, trained using a pretrained weight from a placenta dataset, and tested on 4,455 images, with accuracy, recall, specificity, and F1-score of 99\%, 96\%, 99\%, and 95\% respectively.


Some studies extended the classification of standard planes to ultrasound video clips. In \cite{chen2017ultrasound}, a DL framework was proposed to detect three fetal standard planes in 331 videos, using a Long Short-Term Memory (LSTM) to process temporal information. The accuracy, precision, recall, and F1-score reached 87\%, 71\%, 64\%, and 64\%, respectively. In \cite{pu2021automatic}, a classification CNN and a Recurrent Neural Network (RNN) were used to detect four fetal standard planes in 224 videos, with accuracy, precision, recall, and F1-score of 85\%, 85\%, 85\%, and 85\%, respectively. {\color{red}\cite{slimani2023fetal} introduced an end-to-end fetal biometry and amniotic fluid volume assessment tested on video clips from 172 subjects achieving 95\% agreement between the model and practitioners.}

Different DL detection approaches have been proposed for fetal heart and its internal structure. In one study \cite{dong2019arvbnet}, an SSD model with residual visual blocks was used to detect various heart structures in four-chamber (4CH) images, resulting in an mAP of 93\%. Another study proposed a cardiac-structure localization algorithm \cite{patra2019multi} using a modified VGG-16 and a Faster-RCNN with LSTM layers, resulting in an accuracy of 82\%. In another study, an RNN \cite{huang2017temporal} was used to predict the fetal heart's presence, viewing plane, location, and orientation, resulting in an accuracy of 83\% in correctly classifying views and 79\% in localizing structures. Another study used an end-to-end two-stream full CNN to learn spatio-temporal representations of the fetal heart identification, resulting in accuracy, precision, and recall of 90\%, 85\%, and 89\%, respectively.

Several DL methods in fetal brain analysis focus on structure segmentation using encoder-decoder architectures. \cite{wang2018deep} uses a DL architecture to segment the middle cerebral artery on Doppler ultrasound images and obtains a Dice score of 77\%, Intersection over Union of 63\%, and Hausdorff distance of 26.40 mm. \cite{wu2020automatic} uses a deep attention network to segment the cavum septum pellucidum and get a precision of 79\%, recall of 74\%, Dice score of 77\%, and Hausdorff distance of 0.78 mm on a dataset of 448 ultrasound images. \cite{singh2021semantic} employs ResU-Net to segment the cerebellum and achieves a Dice score of 87\%, Hausdorff distance of 28.15 mm, recall of 86\%, and precision of 90\% on 734 ultrasound images. \cite{zhang2020multiple} uses MA-Net for fetal head circumference segmentation and gets a Dice score of 97\%, a precision of 97\%, a recall of 98\%, and a Hausdorff distance of 10.92 mm on 70 images. {\color{red} \cite{zeng2021fetal} introduced a deeply supervised attention-gated within V-Net backbone achieving a Dice score of 98\%, and a Hausdorff distance of 1.29 mm on 355 fetal head circumference segmentation.}

However, the field of fetal ultrasound analysis faces several challenges, such as the scarcity of publicly available fetal ultrasound imaging datasets, which limits researchers' ability to train and validate their models.  Due to that, researchers often use different datasets to train and evaluate their models. This makes it difficult to compare results between different studies and establish a benchmark for performance. {\color{red} Privacy concerns are the main attribute of not having a public dataset that prevents us from sharing our data. However, instead of performing manual labeling, our method will help label new datasets if someone wants to introduce a new dataset by introducing a clustering method that has not been researched before in the fetal ultrasound images domain.}

\section*{Materials and Methods}
\subsection*{Fetal Ultrasound Scans} We have extracted an unlabeled dataset (88,063 images from 5,425 subjects) from Al Corniche Hospital in Abu Dhabi. The hospital follows the International Society of Ultrasound in Obstetrics and Gynecology (ISUOG) \cite{salomon2011practice} guidelines for fetal ultrasound image acquisition. The dataset received IRB approval. The dataset is anonymized and consists of second-trimester fetal ultrasound scans from one calendar year captured using {\color{red} GE Voluson E8, GE Voluson  E10, GE Voluson S10 Expert, GE Voluson P8, and Philips iU22 machines}. 
\subsection*{Pseudo-labeling} 
\begin{figure*}[th!]
     \centering
         \centering
         \includegraphics[width=0.65\textwidth]{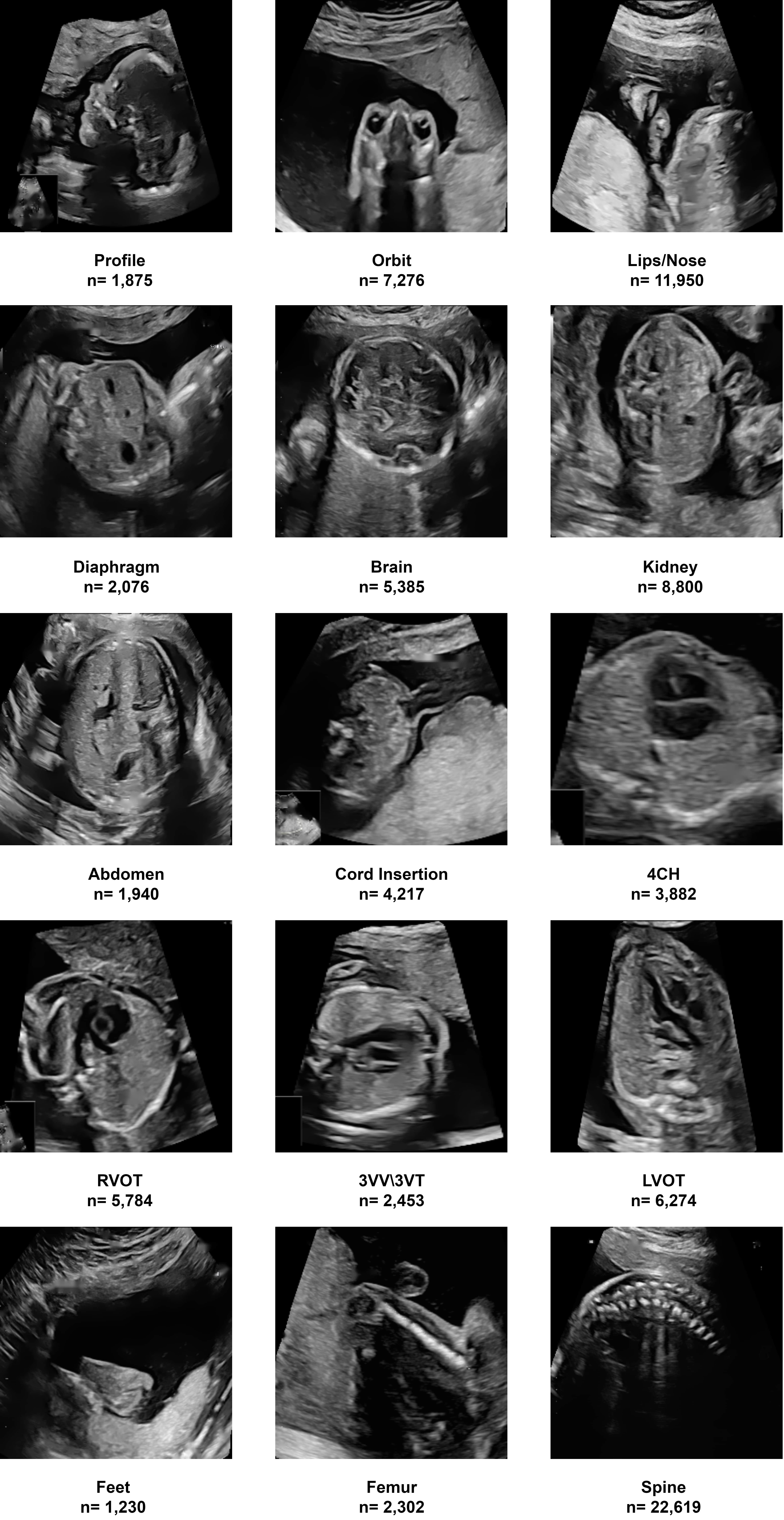}
         \caption{Samples of the views in the dataset, where \emph{n} represents the number of samples.}
         \label{fig:images_samples}
\end{figure*}

To evaluate the clustering algorithm, we have extracted the pseudo labels from the text burnt on the image written by the clinician during the acquisition of scans. We have applied optical character recognition (OCR) with EasyOCR 
to extract the text that identifies the image label. Figure \ref{fig:images_samples} displays examples and a number of samples from our imbalanced dataset. Several views are semantically similar (e.g., heart views such as RVOT, LVOT, and 3VV), which is much more challenging than we typically encounter in natural imaging. We manually reviewed 10\% of labels to ensure high quality pseudo-labels and confirm that the noise level is less than 2\%.


\subsection*{Data Pre-Processing}

To avoid the model learning from the sonographer's texts and only focus on learning features of the fetal organs, we inpaint the text from the ultrasound images following an approach described in \cite{dadoun2021combining}. In addition, we train a convolutional neural network (ResNet18 \cite{he2016deep}) classifier to verify the inpainting process by classifying the views based on the pseudo-labels. We assess via Grad-CAM \cite{selvaraju2017grad} the regions that the network pays attention to when making the classification. When visually reviewing Grad-CAM results on a random subset of the testing set, we observe that the network is paying attention to the fetal organs rather than the inpainted text.

 \subsection*{FUSC: Fetal Ultrasound Semantic Clustering}

\begin{figure*}[th!]
     \centering
         \centering
         \includegraphics[width=1\textwidth]{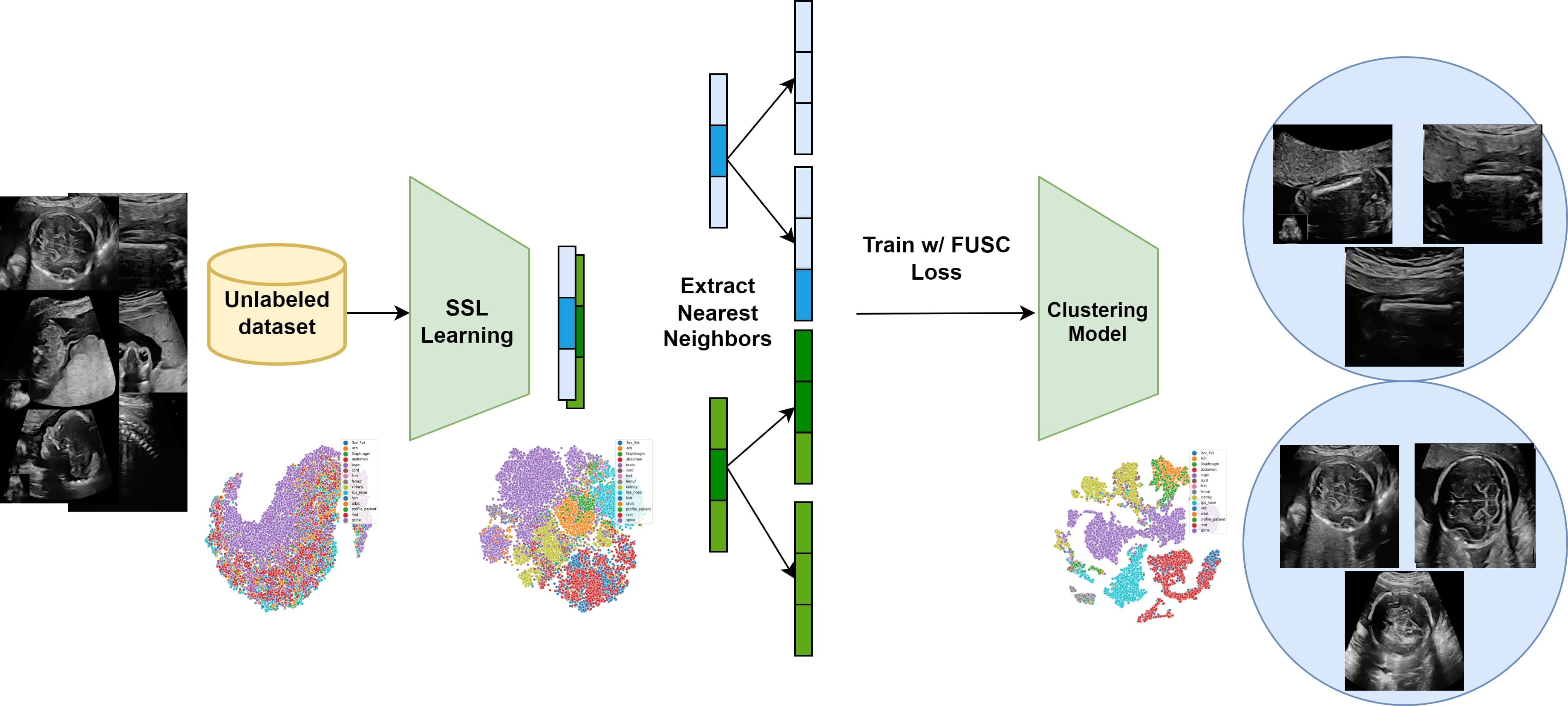}
         \caption{ This framework uses SSL to learn good representation. The nearest neighbors for each image in the embedding space are found. The SSL output trains a clustering model to categorize image embeddings. The blue and green vectors represent different image embedding, whereas we expect embedding with a similar color to represent images from the same class.}
         \label{fig:model framework}
\end{figure*}

We propose a clustering method that aims to disentangle fetal ultrasound views in an unsupervised approach, as shown in Figure \ref{fig:model framework}. The goal of clustering is to generate high cluster purity, which can reduce the time needed for labeling by checking the images that do not belong to the cluster. Our method is inspired by \cite{van2020scan}, which has been demonstrated to work well on natural images. The key steps in our method are: (1) train a SSL network to learn a good fetal view representation as a preliminary step for semantic clustering; (2) introduce a loss function to cluster images and their nearest neighbors into the same category; and (3) propose a self-labeled classification approach to train a network by generating labels from the high confident samples in the clustering model.


\subsubsection*{Self-supervised Learning}
SSL involves training a model $\Phi_\theta$ with parameters $\theta$ using a pretext task $\tau$. One of the primary objectives of SSL is to create an embedding that captures essential low to high dimensional features. The learned weights $\theta$ depend on the chosen pretext task $\tau$. The goal is to develop an embedding where similar images are mapped closely together, which is helpful for clustering. As a result, the selected pretext task should satisfy the requirement of minimizing the distance between image $X_i$ and its augmentation $T[X_i]$. Mathematically, this can be expressed as:
$$
\min_\theta d(\Phi_\theta(X_i), \Phi_\theta(T[X_i])).
$$
Wang and Isola \cite{wang2020understanding} have demonstrated that contrastive learning can minimize this criterion effectively. In our work, we will focus on using two different contrastive learning frameworks SimCLR \cite{chen2020simple} and DINO \cite{caron2021emerging} {\color{red} to illustrate that our framework works with different settings and is not optimized to one specific algorithm. Also, we have used the default backbones family that was used in original papers of SimCLR (ResNet \cite{he2016deep}) and DINO (ViT \cite{dosovitskiy2020image}).
 SimCLR \cite{chen2020simple} is a contrastive SSL method that is based on training the model with positive and negative samples. The goal is to maximize the similarity between positive pairs and minimize the similarity between negative pairs. The positive pairs were extracted by augmentation while the negative pairs were randomly picked from the dataset.
DINO \cite{caron2021emerging} is a self-distillation SSL method based on training teacher-student model instantaneously. The teacher model guides the student network and updates its weight through a gradient of the student model using an exponential moving average of the student parameters. Like SimCLR, the model training is based on positive and negative samples.}




\subsubsection*{Clustering with FUSC loss}
After completing the SSL step, we find each sample's nearest neighbors in the embedding space. This information is then used to train a clustering function $\Phi_\eta$, consisting of a linear layer, that performs clustering on the latent representation of the sample. The loss function in this step is designed to enforce sample $X_i$ and its nearest neighbors $N_{X_i}$ to be assigned to the same cluster. This ensures that semantically similar images are grouped accurately. The function $\Phi_\eta$ utilizes a softmax operation to allocate the sample $X_i$ to different clusters $\mathcal{C} = {1,\ldots,c}$. The output of $\Phi_\eta$ for a sample $X_i$ is a probability distribution in the form of $\Phi_\eta \left(X_i\right) \in [0,1]^C$, with $\Phi_\eta^c(X_i)$ representing the probability of $X_i$ being assigned to cluster $c$. Our objective is to minimize the following:
\begin{equation}
\label{eq:loss_objective}
\begin{split}
\Lambda = -\frac{1}{|\mathcal{D}|}\sum\limits_{X\in\mathcal{D}}\sum\limits_{k\in\mathcal{N}_{X}}&\log\left<\Phi_\eta(X),\Phi_\eta(k)\right> + \lambda\sum_{c\in\mathcal{C}} \Phi_\eta'^c \log \Phi_\eta'^c, \\
&\text{with~} \Phi_\eta'^c = \frac{1}{|\mathcal{D}|}\sum\limits_{X\in\mathcal{D}}\Phi_\eta^c(X).
\end{split}
\end{equation}
{\color{red} ${\mathcal{D}}$ represents the dataset. The second term is an entropy acting as a regularizer to prevent the model from collapsing into a single cluster by promoting a uniform distribution of predictions across different clusters $\mathcal{C}$. The $\lambda$ represents the weight of the uniformity loss. The dot product operation represented by $\left<\cdot\right>$ means the first term of the objective ensures consistent predictions between a sample $X_i$ and its neighboring samples $\mathcal{N}_{X_i}$. The dot product is maximized when the predictions are confident and assigned to the same cluster.}
\subsubsection*{Self-labeling Classification}
It has been shown in \cite{van2020scan} that samples with highly confident predictions ($p_{max} \approx 0.99$) tend to be correctly assigned to their respective clusters. The presence of false positive samples can result from them being near samples from different clusters, causing the network to make uncertain predictions. We propose a self-labeling step in which we assign labels to highly confident samples ($p_{max} \geq threshold = 0.99$) to the cluster they belong to. We then re-train the network as a classification task using a cross-entropy loss applied to strongly augmented confident samples to avoid overfitting. This allows the network to become more confident in its predictions, and more samples can gradually be incorporated into the training process. 

\section*{Experiments \& Results}
\subsection*{Evaluation Metrics}
In order to evaluate the effectiveness of our clustering models, we utilized two metrics: normalized mutual information (NMI) \cite{manning2008introduction}, and cluster purity (CP) \cite{manning2008introduction}. The equations of NMI and CP are as the following:

  \begin{equation}
    NMI(X, Y) = \frac{2I(X;Y)}{H(X) + H(Y)}
    \label{eq:nmi}
  \end{equation}\break
  \begin{equation}
    CP(X, L) = \frac{1}{|\mathcal{D}|}\displaystyle\sum\limits_{c}\max_{j}|{x^c\cap l^j|}
     \label{eq:cp}
  \end{equation}
where $I$ represents the mutual information between sets $X$ and $Y$, $H$ represents the entropy, ${|\mathcal{D}|}$ is the number of images, $x^c$ represents the sample assigned to cluster c, and $l \in L$ represents the ground truth label $l$ from a set $L$.
\subsection*{Experimental setup}
Typically, clustering is performed on the complete dataset \cite{ji2019invariant,caron2018deep,kart2021deepmcat}. In our case, we divided the data {\color{red} randomly} by subject and view into an 80\% training set and a 20\% testing set to evaluate the model's generalizability on unseen data. We use 70,446 images for training and 17,617 for testing. There is no need for a validation set in clustering models because we assume no labels exist in the training. We have utilized two SSL methods, SimCLR \cite{chen2020simple}, and DINO \cite{caron2021emerging}. For the clustering step, {\color{red}We chose the top twenty nearest neighbors}, as the model is not sensitive to this hyperparameter \cite{van2020scan}. We also employed the same hyperparameter settings for the SCAN model outlined in \cite{van2020scan}, such as $\lambda = 5$, because it experimentally shows the best results. 
\subsection*{Results}
\begin{figure}[ht]
    \centering
    \begin{subfigure}{0.2\textwidth}
        \includegraphics[width=0.9\textwidth]{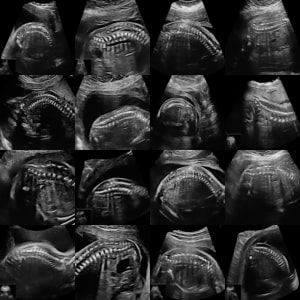}
        \label{fig:cluster_0}
    \end{subfigure}%
    \hfill
    \begin{subfigure}{0.2\textwidth}
        \includegraphics[width=0.9\textwidth]{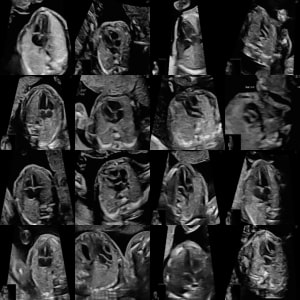}
        \label{fig:cluster_1}
    \end{subfigure}%
    \hfill
    \begin{subfigure}{0.2\textwidth}
        \includegraphics[width=0.9\textwidth]{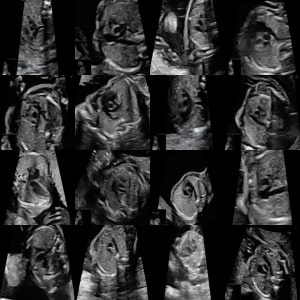}
    \end{subfigure}%
    \hfill
    \begin{subfigure}{0.2\textwidth}
        \includegraphics[width=0.9\textwidth]{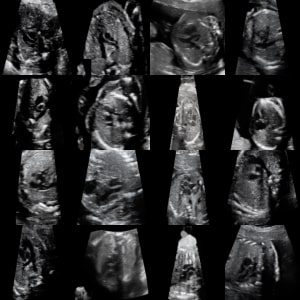}
    \end{subfigure}%
    \hfill
    \begin{subfigure}{0.2\textwidth}
        \includegraphics[width=0.9\textwidth]{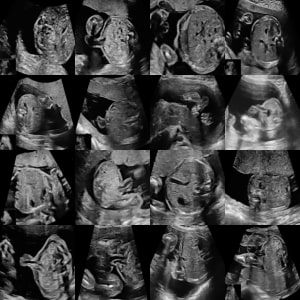}
    \end{subfigure}%
    \hfill
    \begin{subfigure}{0.2\textwidth}
        \includegraphics[width=0.9\textwidth]{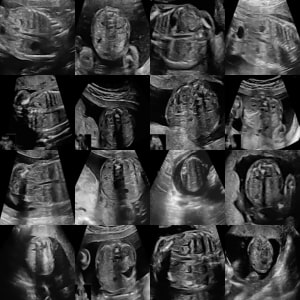}
    \end{subfigure}%
    \hfill
    \begin{subfigure}{0.2\textwidth}
        \includegraphics[width=0.9\textwidth]{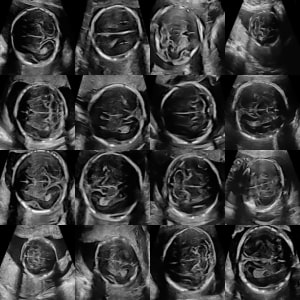}
    \end{subfigure}%
    \hfill
    \begin{subfigure}{0.2\textwidth}
        \includegraphics[width=0.9\textwidth]{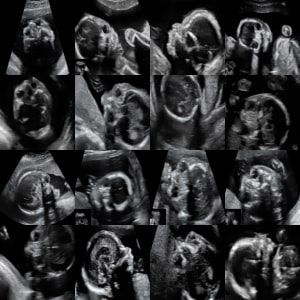}
    \end{subfigure}%
    \hfill
    \begin{subfigure}{0.2\textwidth}
        \includegraphics[width=0.9\textwidth]{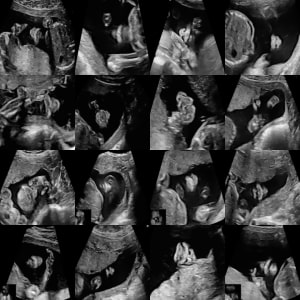}
    \end{subfigure}%
    \hfill
    \begin{subfigure}{0.2\textwidth}
        \includegraphics[width=0.9\textwidth]{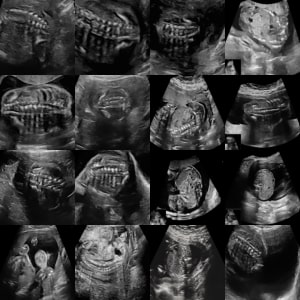}
    \end{subfigure}%
    \hfill
    \begin{subfigure}{0.2\textwidth}
        \includegraphics[width=0.9\textwidth]{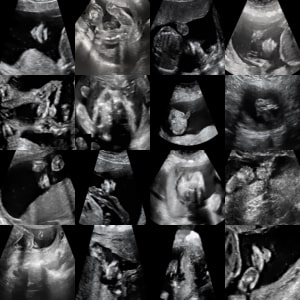}
    \end{subfigure}%
    \hfill
    \begin{subfigure}{0.2\textwidth}
        \includegraphics[width=0.9\textwidth]{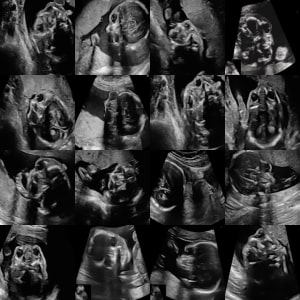}
    \end{subfigure}%
    \hfill
    \begin{subfigure}{0.2\textwidth}
        \includegraphics[width=0.9\textwidth]{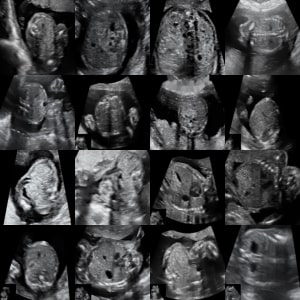}
    \end{subfigure}%
    \hfill
    \begin{subfigure}{0.2\textwidth}
        \includegraphics[width=0.9\textwidth]{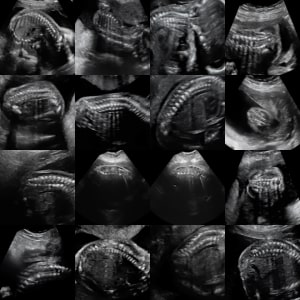}
    \end{subfigure}%
    \hfill
    \begin{subfigure}{0.2\textwidth}
        \includegraphics[width=0.9\textwidth]{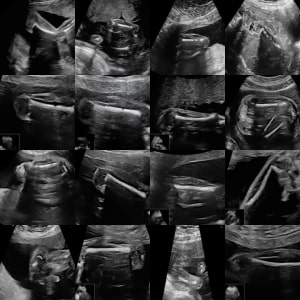}
    \end{subfigure}%
    \hfill
    \caption{Samples of clusters of $FUSC^*_{simCLR}$ model.}
    \label{fig:all_clusters}
\end{figure}

We summarize our findings in Tables \ref{results_table}-\ref{tab:CP-one-class}. Initially, we re-implemented \cite{kart2021deepmcat} by modifying the DeepCluster framework \cite{caron2018deep} using the same configuration they employed. Although the work in \cite{kart2021deepmcat} has shown impressive results in clustering the well-define problem of cardiac MR images, it provides the lowest performance when clustering fetal ultrasound view. When K-means is applied to the embedding space, it has demonstrated better results compared to \cite{kart2021deepmcat}. Moreover, the FUSC model's performance was superior to \cite{kart2021deepmcat} and K-means. Furthermore, self-labeling improved the model's performance (tagged with $^*$ in Table \ref{results_table}). We included the results of supervised learning in determining the model's upper limit, and there is a 21\% gap in the CP. {\color{red} Figure \ref{fig:all_clusters} shows images assigned to the same cluster by our $FUSC^*_{SimCLR }$ model.}

\subsection*{Ablation Studies}
\subsubsection*{Clustering By Merging Semantically Similar Views}
To investigate the impact of clustering with merging semantically similar views, we combine classes into four categories: Heart (RVOT, LVOT, 4CH, and 3VV/3VT), Head (Brain, Profile, Orbit, and Lips/Nose), Abdomen (Abdomen, Kidney, Diaphragm, and Cord Insertion), and Bone (Spine, Feet, and Femur). The clustering result is presented in {\color{red}Table \ref{tab:under_clustering}, indicating having an equivalence or increasing in the CP compared to the 15 classes. }


 \subsubsection*{Over Clustering}
 Following the approach of \cite{kart2021deepmcat}, where the number of clusters were eight times greater than the number of classes, we evaluated whether this technique could enhance our model's performance by clustering the data into 120 clusters. The results are summarized in Table \ref{tab:over clustering}. Initially, we observed increased CP compared to using only 15 clusters in all the models. Furthermore, the self-labeling step resulted in lower CP and fewer filled clusters while having higher NMI.

\subsubsection*{Model Generalizability}
To test the generalizability of our clustering model, we used an additional dataset for testing purposes only. It contains publicly available fetal views ultrasound images of six classes, including the abdomen, brain, maternal cervix, femur, thorax, and others captured using {\color{red} Aloka, GE Voluson E6, and GE Voluson S10 machines} \cite{burgos2020evaluation}.  We encounter a distribution shift as ultrasound machines are different, and there are out-of-distribution classes, such as the maternal cervix. To ensure consistency, we applied the same preprocessing as described above and excluded the other class. We utilized the $FUSC^*_{SimCLR}$ weights with 15 clusters of our pre-trained model to perform clustering without any fine-tuning. The outcomes demonstrate that our model achieves a high CP rate of 92\% and an NMI of 72\%.

\section*{Discussion \& Conclusion}
 We show in Table \ref{results_table} that FUSC outperforms different clustering methods, especially the work presented in \cite{kart2021deepmcat}, achieving 72\% CP and 68\% NMI. However, a 21\% gap in CP compared to supervised training is observed. We have reached this performance by training a model without any labels, and  we attribute this large gap due to the challenging problem at hand, making it difficult to distinguish some views without human guidance.
 
\begin{figure*}[th!]
     \centering
     \centering
     \includegraphics[width=0.75\textwidth]{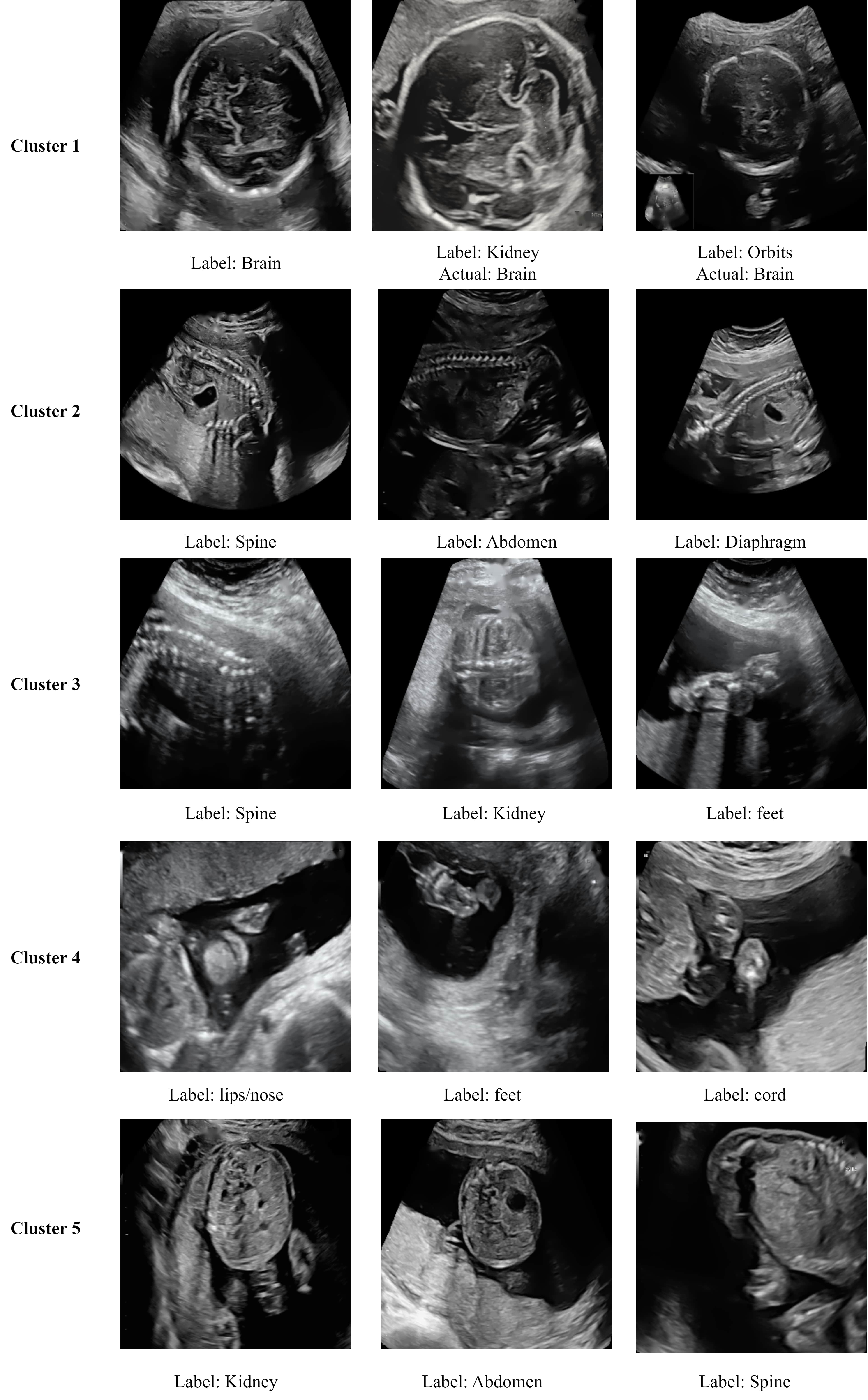}
     \caption{Samples of images in the best five clusters for the $FUSC^*_{simCLR}$ model.}
     \label{fig:cluster_images}
\end{figure*}

 Table \ref{tab:CP-one-class} presents the top class within each cluster of $FUSC^*_{simCLR}$ model. The top five clusters show high CP above 95\%, indicating the effectiveness of our model. Figure \ref{fig:cluster_images} illustrates samples in the top five clusters. By inspecting them, we found that the main reasons for {\color{red} lower performance in certain clusters. Despite accurate clustering in cluster 1, the ground truth was mislabeled, leading to underscores the performance. Cluster 2 presented challenges due to the inclusion of views containing multiple structures, such as Abdomen and Spine or Diaphragm and Spine. Clusters 3 and 4 contain images that have a similar overall appearance.}. When assessing the most entangled clusters, we observe that they contain images from semantically similar views, which also typically confuse clinicians. For example, the lowest cluster contains RVOT, LVOT, and 4CH as the most dominant classes, all belonging to heart views. Although the entropy-based loss should help with the class imbalance issue, we believe it still contributes to a lower CP in some clusters. We observe that the most frequent classes dominate multiple clusters, e.g., images from the spine view dominate three clusters (2, 3, and 8).

By merging semantically similar views, we observe {\color{red}a higher or equivalence CP than clustering 15 classes illustrated in Table \ref{tab:under_clustering})}. This emphasizes the model's ability to discriminate views that are less semantically similar, where the biggest problem is from views that are challenging to distinguish. {\color{red} However, it is worth noting that the self-labeling step may degrade the clustering, potentially resulting in a lower CP. Nonetheless, the model's performance remains comparable to the clustering of 15 classes.}

We show in Table \ref{tab:over clustering} that over-clustering has shown higher CP. We believe this is because of the increase in the number of clusters where the NMI was lower in the 15 clusters setup compared to over-clustering. Also, the self-labeling has reduced the number of the filled clusters. {\color{red}The reason is in the self-labeling step uses a cross-entropy loss only which does not encourage having a uniform number of samples in each cluster. In contrast, the FUSC loss has a regularizer that encourages having a uniform number of samples in each cluster.}

Our model shows good generalizability when we tested on an additional unseen dataset that consists of distribution shift and out-of-distribution samples. Our CP reached 92\%, which is higher than all the baselines we compare within our dataset because the views in that dataset do not have strong semantic relations views.
We present a fetal ultrasound view self-supervised clustering method and evaluate its performance with extracted pseudo labels from the images. We build our method on a large dataset and demonstrate its generalizability by testing on an additional set. Additionally, we had a quantitative and qualitative analysis per cluster to have a better understanding of the failed cases. We found that the model performance degraded due to the imbalance of the dataset and views that have high semantic similarity.  As much as we attempted to bridge the gap between the ability of clustering and supervised classification to label fetal ultrasound images, this is still an open research problem.

In future work, we aim to enhance the model performance by reducing the effect of the imbalance of the dataset. This imbalance can pose a challenge in accurately clustering fetal ultrasound images, particularly when certain views are underrepresented. Additionally, we plan to explore advanced techniques for feature extraction and representation learning to improve the disentanglement of semantically similar views, which will further refine the clustering process. By doing so, we hope to contribute to the ongoing efforts to make fetal ultrasound image analysis more accurate and reliable, ultimately benefiting the field of prenatal healthcare and diagnostics.

\section*{Conflict of Interest Statement}
The authors declare no competing interests.

\section*{Data Availability Statement}
The raw/processed data required to reproduce the above findings cannot be shared at this time due to legal/ ethical reasons.

\section*{Declaration of Generative AI}
During the preparation of this work, the authors used ChatGPT in order to enhance writing. After using this tool/service, the authors reviewed and edited the content as needed and take full responsibility for the content of the publication.




\bibliographystyle{elsarticle-num}
\bibliography{keylatex.bib}






\begin{table*}[ht!]
\centering
\caption{Results of applying a variant of models in the dataset. ($^*$) refers to the self-labeling step.}
\label{results_table}
\begin{tabular}{|c|c|c|c|c|}
\hline
Model             & Backbone   & Number of Clusters & CP & NMI  \\ \hline
DeepMCAT \cite{kart2021deepmcat} & VGG16  & 15 & 35\% & 16\% \\ \hline
K-mean + SimCLR    & ResNet18         & 15 & 48\%           & 33\% \\ \hline
K-mean + DINO     & ViT-S/16       & 15 & 69\%           & 62\% \\ \hline
$FUSC_{SimCLR }$     &  ResNet18 & 15 & 64\%           & 55\% \\ \hline
$FUSC^*_{SimCLR }$ &  ResNet18 & 15 & 71\%           & 65\% \\ \hline
$FUSC_{Dino}$             &  ViT-S/16 & 15 & 71\%           & 67\% \\ \hline
$FUSC^*_{Dino}$             &  ViT-S/16 & 15 & \textbf{72\%}           & \textbf{68\%} \\ \hline
Supervised & ResNet18  & 15 & 93\% & 86\% \\ \hline
\end{tabular}
\end{table*}

\begin{table}[ht]
\centering
\caption{Result of clustering by merging semantically similar views. ($^*$) refers to the self-labeling step.}
\label{tab:under_clustering}
\begin{tabular}{|c|c|c|c|c|}
\hline
Model             & Backbone & Number of Clusters & CP   & NMI  \\ \hline
$FUSC_{SimCLR }$  & ResNet18 & 4                  & \textbf{87\%} & \textbf{70\%} \\ \hline
$FUSC^*_{SimCLR }$& ResNet18 & 4                  & 76\% & 64\% \\ \hline
$FUSC_{Dino}$     & ViT-S/16 & 4                  & 72\% & 48\% \\ \hline
$FUSC^*_{Dino}$   & ViT-S/16 & 4                  & 64\% & 33\% \\ \hline
\end{tabular}
\end{table}

 \begin{table*}[ht]
\centering
\caption{Results of applying over clustering. We notice that self-labeling negatively affects cluster purity since it may lead to fewer filled clusters. ($^*$) refers to the self-labeling step.}
\label{tab:over clustering}
\begin{tabular}{|c|c|c|c|c|}
\hline
Model             & Backbone        & Filled clusters & CP & NMI  \\ \hline
DeepMCAT \cite{kart2021deepmcat} & VGG16 &  120& 44\% & 22\% \\ \hline
$FUSC_{SimCLR }$          & ResNet18 & 120                & 76\%           & 49\% \\ \hline
$FUSC^*_{SimCLR }$ &  ResNet18 & 35               & 72\%           & 57\% \\ \hline
$FUSC_{Dino}$           & ViT-S/16 & 120             & \textbf{82\%}           & 55\% \\ \hline
$FUSC^*_{Dino}$           & ViT-S/16 & 15                & 68\%           & \textbf{63}\% \\ \hline
\end{tabular}
\end{table*}

\begin{table*}[th!]
\centering
\caption{Overview of the top three classes in each cluster and the percentage of them for the $FUSC^*_{simCLR}$ model.}
\label{tab:CP-one-class}
\begin{tabular}{|c|cc|cc|cc|c|}
\hline
\textbf{Cluster} & \multicolumn{2}{c|}{Top Class 1}                      & \multicolumn{2}{c|}{Top Class 2}      & \multicolumn{2}{c|}{Top Class 3}                     & Cluster Size \\ \hline
\textbf{1}       & \multicolumn{1}{c|}{Brain}                    & 100\% & \multicolumn{1}{c|}{Orbit}     & 0\%  & \multicolumn{1}{c|}{Kidney}                   & 0\%  & 914          \\ \hline
\textbf{2}       & \multicolumn{1}{c|}{Spine}                    & 99\%  & \multicolumn{1}{c|}{Diaphragm} & 0\%  & \multicolumn{1}{c|}{Lips\textbackslash{}Nose} & 0\%  & 2,416        \\ \hline
\textbf{3}       & \multicolumn{1}{c|}{Spine}                    & 97\%  & \multicolumn{1}{c|}{Kidney}    & 1\%  & \multicolumn{1}{c|}{Diaphragm}                & 1\%  & 402          \\ \hline
\textbf{4}       & \multicolumn{1}{c|}{Lips\textbackslash{}Nose} & 97\%  & \multicolumn{1}{c|}{Feet}      & 2\%  & \multicolumn{1}{c|}{Cord Insertion}           & 0\%  & 1,955        \\ \hline
\textbf{5}       & \multicolumn{1}{c|}{Kidney}                   & 95\%  & \multicolumn{1}{c|}{Abdomen}   & 2\%  & \multicolumn{1}{c|}{Spine}                    & 1\%  & 988          \\ \hline
\textbf{6}       & \multicolumn{1}{c|}{Femur}                    & 89\%  & \multicolumn{1}{c|}{Feet}      & 5\%  & \multicolumn{1}{c|}{Spine}                    & 2\%  & 498          \\ \hline
\textbf{7}       & \multicolumn{1}{c|}{Orbit}                    & 79\%  & \multicolumn{1}{c|}{Profile}   & 14\% & \multicolumn{1}{c|}{Brain}                    & 5\%  & 1,690        \\ \hline
\textbf{8}       & \multicolumn{1}{c|}{Spine}                    & 68\%  & \multicolumn{1}{c|}{Diaphragm} & 7\%  & \multicolumn{1}{c|}{Kidney}                   & 6\%  & 2,427        \\ \hline
\textbf{9}       & \multicolumn{1}{c|}{Orbit}                    & 53\%  & \multicolumn{1}{c|}{Profile}   & 25\% & \multicolumn{1}{c|}{Brain}                    & 22\% & 105          \\ \hline
\textbf{10}      & \multicolumn{1}{c|}{Kidney}                   & 52\%  & \multicolumn{1}{c|}{Diaphragm} & 18\% & \multicolumn{1}{c|}{Cord Insertion}           & 14\% & 407          \\ \hline
\textbf{11}      & \multicolumn{1}{c|}{Lips\textbackslash{}Nose} & 47\%  & \multicolumn{1}{c|}{Feet}      & 19\% & \multicolumn{1}{c|}{Profile}                  & 13\% & 622          \\ \hline
\textbf{12}      & \multicolumn{1}{c|}{LVOT}                     & 44\%  & \multicolumn{1}{c|}{4CH}       & 30\% & \multicolumn{1}{c|}{RVOT}                     & 23\% & 1,330        \\ \hline
\textbf{13}      & \multicolumn{1}{c|}{RVOT}                     & 39\%  & \multicolumn{1}{c|}{LVOT}      & 29\% & \multicolumn{1}{c|}{4CH}                      & 24\% & 410          \\ \hline
\textbf{14}      & \multicolumn{1}{c|}{Cord Insertion}           & 39\%  & \multicolumn{1}{c|}{Kidney}    & 28\% & \multicolumn{1}{c|}{Abdomen}                  & 21\% & 1,503        \\ \hline
\textbf{15}      & \multicolumn{1}{c|}{RVOT}                     & 35\%  & \multicolumn{1}{c|}{LVOT}      & 27\% & \multicolumn{1}{c|}{3VV\textbackslash{}3VT}   & 22\% & 1,950        \\ \hline
\end{tabular}
\end{table*}
\end{document}